\documentclass[letterpaper]{article} % DO NOT CHANGE THIS
\usepackage{aaai24}  % DO NOT CHANGE THIS
\usepackage{times}  % DO NOT CHANGE THIS
\usepackage{helvet}  % DO NOT CHANGE THIS
\usepackage{courier}  % DO NOT CHANGE THIS
\usepackage[hyphens]{url}  % DO NOT CHANGE THIS
\usepackage{graphicx} % DO NOT CHANGE THIS
\usepackage{natbib}  % DO NOT CHANGE THIS AND DO NOT ADD ANY OPTIONS TO IT
\usepackage{caption} % DO NOT CHANGE THIS AND DO NOT ADD ANY OPTIONS TO IT
\usepackage{algorithm}
\usepackage{algorithmic}
\usepackage{multirow}
\usepackage{algorithm}
\usepackage{algorithmic}
\usepackage{amsmath}
\usepackage{amssymb}
\usepackage{stackengine}
\usepackage{rotating}
\usepackage{xcolor}
\usepackage{array}
\usepackage{booktabs}
\usepackage{newfloat}
\usepackage{listings}
\DeclareCaptionStyle{ruled}{labelfont=normalfont,labelsep=colon,strut=off} % DO NOT CHANGE THIS
\floatstyle{ruled}
\newfloat{listing}{tb}{lst}{}
\floatname{listing}{Listing}
\title{PoseGen: Learning to Generate 3D Human Pose Dataset with NeRF}
\author {
    Mohsen Gholami,
    Rabab Ward,
    Z. Jane Wang
}
\affiliations {
    University of British Columbia, Vancouver, Canada\\
    \{mgholami, rababw, zjanew\}@ece.ubc.ca
}

\usepackage{bibentry}
\begin{document}

\maketitle

%%%%%%%%% ABSTRACT
\begin{abstract}
This paper proposes an end-to-end framework for generating 3D human pose datasets using Neural Radiance Fields (NeRF). Public datasets generally have limited diversity in terms of human poses and camera viewpoints, largely due to the resource-intensive nature of collecting 3D human pose data. As a result, pose estimators trained on public datasets significantly underperform when applied to unseen out-of-distribution samples. Previous works proposed augmenting public datasets by generating 2D-3D pose pairs or rendering a large amount of random data. Such approaches either overlook image rendering or result in suboptimal datasets for pre-trained models. Here we propose \emph{PoseGen}, which learns to generate a dataset (human 3D poses and images) with a feedback loss from a given pre-trained pose estimator. In contrast to prior art, our generated data is optimized to improve the robustness of the pre-trained model. 
The objective of \emph{PoseGen} is to learn a distribution of data that maximizes the prediction error of a given pre-trained model. As the learned data distribution contains OOD samples of the pre-trained model, sampling data from such a distribution for further fine-tuning a pre-trained model improves the generalizability of the model. This is the first work that proposes NeRFs for 3D human data generation. NeRFs are data-driven and do not require 3D scans of humans. Therefore, using NeRF for data generation is a new direction for convenient user-specific data generation. Our extensive experiments show that the proposed \emph{PoseGen} improves two baseline models (SPIN and HybrIK) on four datasets with an average 6\% relative improvement. 
\end{abstract}

%%%%%%%%% BODY TEXT
\section{Introduction}
\begin{figure}[t]
\begin{center}
   \includegraphics[width=1\linewidth]{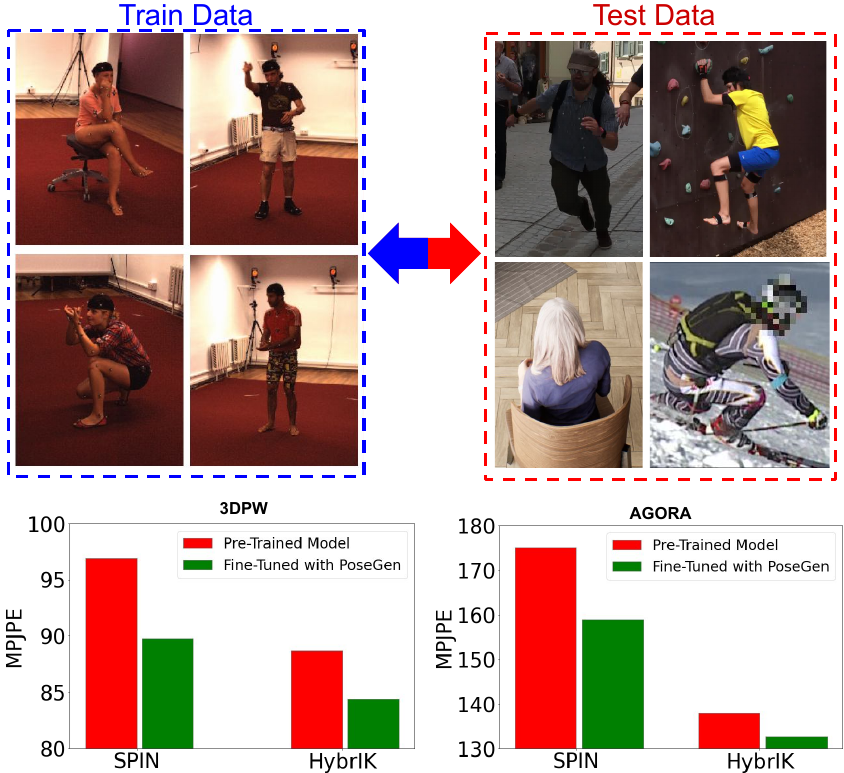}
\end{center}
   \caption{There is a domain gap between the training data used for training pose estimators and in-the-wild images (OOD samples). Therefore, pre-trained models underperform when applied to OOD samples.}
\label{fig:tiser}
\end{figure}

3D human pose and mesh estimation, the task of reconstructing human pose in 3D space given a 2D image of the person, is an ill-posed problem, and many data-driven approaches using deep learning were recently proposed \cite{liu2023poseexaminer,GHOLAMI202297,li2022cliff}. A dataset covering the distribution of possible human poses, global orientation, appearance, and other attributes would be large and difficult to capture in practice. In contrast to vision tasks such as classification and object detection, 3D human pose labels can not be obtained by manual annotation and require an expensive setting for accurate measuring. Therefore, there are limited public datasets and these datasets generally have limited diversity. Unfortunately, most pose estimation models that are trained on public datasets underperform when applied to the out-of-distribution (OOD) samples or samples in the tails of the distribution. Figure \ref{fig:tiser} shows the domain gap between training and test data.

To address the above-mentioned concern, prior work proposed augmenting the publicly available datasets by generating 2D-3D human pose pairs \cite{Gong_2021_CVPR, Gholami_2022_CVPR, Li_2020_CVPR,Li_Pun_2023} or by rendering synthetic images in more variant poses and appearances \cite{agora,bedlam}. 
Augmenting a 3D dataset in the 2D-3D pose space is useful merely for the models that accept 2D poses as input. Therefore, the majority of 3D pose estimators that accept RGB images as input (not 2D pose) can not be fine-tuned with augmented 2D-3D pose pairs. 
AGORA \cite{agora} and BEDLAM \cite{bedlam} generate synthetic human images and 3D poses. Their experiments show that fine-tuning pre-trained pose estimators on synthetic data makes the models robust on OOD test samples.
However, AGORA and BEDLAM use game engines to render human images in an offline manner. Therefore the generated datasets are not optimized for a particular pre-trained model to improve its robustness and generalizability.

Most of the prior arts consider data generation and model training as two different steps (offline methods). 
% Therefore, it is not clear whether generated samples make the model more robust and improve the performance on OOD samples.
PoseAug \cite{Gong_2021_CVPR} and AdaptPose \cite{Gholami_2022_CVPR} proposed data generation and model training in a single step (online methods). They used feedback from a pose estimator to guide data generation. Online methods make data generation a learnable procedure and prevent generating samples that deviate the downstream model from its objective.
PoseAug and AdaptPose use the fixed-hard ratio loss that controls the hardness of generated data. The fixed hard ratio loss converges to zero by either minimizing the loss of the model on the source data or maximizing the loss on the generated data. Such a loss function might converge to generating in-distribution (IND) data. Moreover, fixed hard ratio loss is not source-dataset-free and inevitably needs a source dataset during data generation.
Here we propose directly maximizing the loss of a pre-trained model while generating data to find a distribution of data that is OOD for the pre-trained model.

In this work, we leverage the advances in Neural Radiance Fields (NeRF). NeRF can generate high-quality images of the scene from novel views. 
Compared with traditional rendering engines, NeRF has two major merits: 1) It is differentiable, and 2) it only requires multi-view images for training, and does not require hand-crafted 3D models. Recent works have trained NeRFs on human images and allow the rendering of human images from novel camera viewpoints and novel poses.  NeRFs can be trained on user-specific images and therefore can be used to generate user-specific 3D pose datasets. Having such a dataset significantly improves the accuracy of pre-trained models on tasks that require accurate pose estimation (e.g., medical applications). 

\begin{figure*}[t]
\begin{center}
   \includegraphics[width=0.8\linewidth]{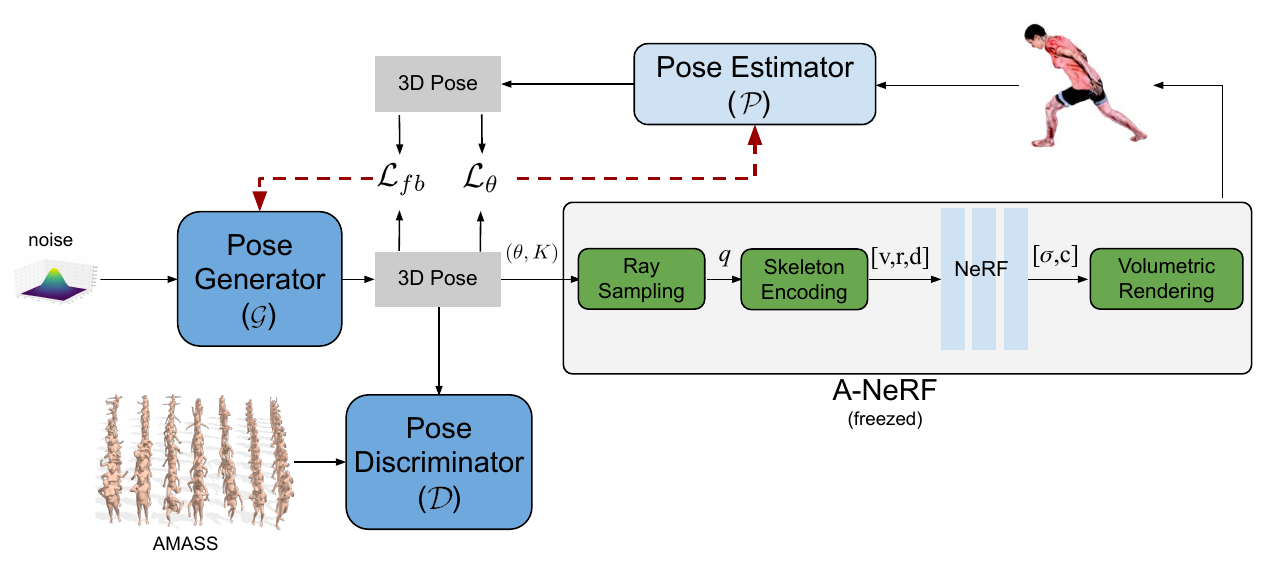}
\end{center}
   \caption{ The overall framework of the proposed \emph{PoseGen}. The pose generator learns to generate input of novel poses $\theta$ and camera viewpoints $K$ which are fed to a NeRF model to render human images. The 3D pose estimator is trained with rendered images and provides feedback to the pose generator. The feedback function enforces the generator to generate OOD data.}
\label{fig:framework}
\end{figure*}

Figure \ref{fig:framework} shows the overall framework of the proposed \emph{PoseGen}. \emph{PoseGen} has a generator that outputs 3D human pose and camera viewpoint. The generated 3D poses are fed to a discriminator that enforces them to be plausible. The generated 3D poses and camera viewpoints are fed to a NeRF model to render corresponding human images. The rendered images are then used to estimate the generated 3D poses. The error of 3D pose estimation is used as feedback to the generator. We investigate two scenarios where we \emph{maximize} or \emph{minimize} the feedback loss from the pose estimator.  Maximizing the loss of pose estimator during data generation leads to OOD data generation and minimizing the loss of pose estimator during the data generation results in IND data generation. 

In summary, our \textbf{contributions} are as follows: We
\begin{itemize}
    \item propose an end-to-end framework for generating novel user-specific 3D human pose and image datasets.
    \item propose a generative model that learns the distribution of a pre-trained model and can generate in-distribution and out-of-distribution poses and images.
    \item propose a simple yet effective feedback function for generative models from pre-trained pose estimators.
    \item show the effectiveness of NeRF for generating human synthetic datasets.
    \item obtain SOTA results when doing extensive experiments on 4 datasets with two baseline models.
\end{itemize}

\section{Related Work}
% In this section, we discuss related work with a focus on human 3D data generation. 

\textbf{Synthetic 2D-3D Pose Generator.} Some prior arts use a two-step method for 3D human pose estimation; In the first step, 2D poses are estimated, and the 3D pose model is trained to estimate 3D from 2D. \cite{Li_2020_CVPR,Gong_2021_CVPR,Gholami_2022_CVPR} propose augmenting 2D-3D pose pairs to improve the robustness of pose estimators that use the two-step method. 
\cite{Li_2020_CVPR} uses an evolutionary method to augment 3D poses by substituting body parts of real poses from a public dataset. 
Their evolutionary method is a random data augmentation without any feedback from the pose estimators.
\cite{Gong_2021_CVPR,Gholami_2022_CVPR} propose a learnable framework that learns how to augment data given feedback loss function from the pose estimator. 
These methods are effective when accurate 2D poses are available at the test time. However, the improvements are limited when 2D pose inputs are in-accurate \cite{Gholami_2022_CVPR}. 

\textbf{Synthetic Image-3D Generator.} Prior works use traditional rendering engines to render photo-realistic human images given human poses. 
AGORA \cite{agora} uses 4240 high-quality textured scans of people and randomly samples 3D people and
place them in scenes at random distances and orientations.
AGORA uses a game engine \cite{unreal} optimized for high-quality output to render human images. 
BEDLAM \cite{bedlam} uses 271 body shapes with 100 skin textures and 27 different types of hair to
the head of SMPL-X. 
Both AGORA and BEDLAM perform data generation in an offline manner and are unable to generate user-specific datasets. 
SURREAL applies primitive textures on naked SMPL body mesh to generate synthetic images. SURREAL \cite{Varol_2017_CVPR} uses 3D sequences of MoCap data and therefore has small variations in terms of body poses. 

There is another direction that uses realistic human images and then renders them synthetically in new scenes \cite{gabeur2019moulding,mehta2017monocular,singleshotmultiperson2018}. 
These methods have the limitation of real data, including a limited variation of human poses. 
Based on realistic human images, \cite{NIPS2016_35464c84} uses 3D pose to select real image whose 2D pose locally matches the projected 3D pose. 
Selected images are then stitched to generate a new synthetic image. 
The generated images by these methods are prone to be unrealistic. Since rendering photo-realistic images is challenging, some prior work rendered SMPL mesh as a silhouette or body segments and estimated 3D human poses \cite{9010054,9009023,pavlakos2018learning}. These methods do not tackle the performance drops due to unseen human appearances and textures. 

\textbf{Pose Priors.} Our work is related to a line of research that uses a machine-learning model to learn the priors of human 3D poses. VPoser \cite{vposer} proposes a Variational Autoencoder to learn a low dimensional latent space for human 3D poses. The learned priors by VAEs are mean-centered and therefore discard the tails of distribution that are far away from the center of the distribution. On the other hand, since its Gaussian prior is unbounded, it is possible to sample poses very far from the mean of the distribution, leading to implausible data.  \cite{davydov2022adversarial} proposes adversarial training to learn bounded priors that are able to sample poses far away from the center of the distribution. None of these works is able to learn a distribution of poses that are plausible while OOD for a given pre-trained model. Therefore, these previous works are not able to effectively improve the generalizability of pre-trained models.

\section{Method}
\subsection{Problem Formulation}

The overall framework of \emph{PoseGen} is given in Fig \ref{fig:framework}. It includes a pose generator $\mathcal{G}$ that outputs human poses $\theta$ and camera viewpoints $K$, a Discriminator $\mathcal{D}$ 
 that enforces generated poses to be plausible, a NeRF model that renders human images given poses and camera viewpoints, and a pre-trained 3D pose estimator $\mathcal{P}$ that learns the new data and provides feedback to the generator. The overall objective is to make the 3D pose estimator $\mathcal{P}$ generalizable to unseen (OOD) samples.  

The generator samples vector $z$ from a distribution $\mathbb{P}_z \in \mathbb{R}^{D}$. The output of generator is SMPL body poses $\theta \in \mathbb{R}^{69}$ and camera viewpoint $K \in \mathbb{R}^{3}$. $\theta$ is the relative rotation of limbs in a format of the axis-angle rotation matrix and $K$ is the camera viewpoint in an axis-angle rotation format. The body shape parameters of SMPL $\beta$ are kept fixed due to the constraints of A-NeRF in rendering bodies with different shapes. Since we aim to generate samples that improve the generalizability of $\mathcal{P}$, it is critical to learn a distribution of data that includes OOD samples.
% while having a distribution not deviation significantly from the initial distribution. 
However, generating OOD samples might make the model deviate from the ideal performance.
Therefore, we perform two sets of experiments. \textit{Scenario 1:} $\mathbb{P}_z$ is learned to be IND for $\mathcal{P}$, and in \textit{Scenario 2:} $\mathbb{P}_z$ is learned to be OOD for $\mathcal{P}$.

Our method learns the latent space distribution of a generative network $\mathcal{G}$ that outputs the input parameters of a NeRF model $C_\phi (\theta,K)$. In \textit{Scenario 1}, the objective of $\mathcal{G}$ is to generate plausible poses while minimizing the training loss of the pose estimator $\mathcal{P}$:

\begin{equation}
    \text{\stackunder{ min }{$\mathcal{G}$,$\mathcal{P}$}}\text{\stackunder{ max  }{$\mathcal{D}$}} \text{$\mathcal{L}$}(\text{$\mathcal{G}$},\text{$\mathcal{D}$},\text{$\mathcal{P}$}).
\end{equation}
The learned latent distribution $\mathbb{P}$ can be used to generate human data (images and poses) that will be considered as IND for $\mathcal{P}$. In \textit{Scenario 2} the objective of $\mathcal{G}$ is to generate plausible poses while increasing the training loss of the pose estimator $\mathcal{P}$:

\begin{equation}
    \text{\stackunder{ min }{$\mathcal{G}$}}\text{\stackunder{ max  }{$\mathcal{D}$,$\mathcal{P}$}} L(\text{$\mathcal{G}$},\text{$\mathcal{D}$},\text{$\mathcal{P}$}).
\end{equation}
The learned latent distribution $\mathbb{P}$ can be used to generate OOD samples for $\mathcal{P}$. The OOD sample contains a distribution of data that covers failure cases of $\mathcal{P}$ including OOD camera viewpoints and human poses. In the following, we will discuss the formulation of $\mathcal{C}, \mathcal{G}$, and $\mathcal{D}$.

\subsection{Pose Generator and Discriminator}
The pose generator samples vector $z$ from a distribution $\mathbb{P}_z$. 
$\mathbb{P}_z$ is the prior of the synthetic data that plays a crucial role in the quality of generated data. We try different latent space distributions including normal, uniform, and spherical distribution $\mathcal{S}$, for data generation:
\begin{align}
z_N\sim\mathbb{P}_z&=\mathcal{N}(0,1)\subset\mathbb{R}^{d}, \\
z_U\sim\mathbb{P}_z&=\mathcal{U}_{ [-1,1]^{d}}\subset\mathbb{R}^{d}, \\
z_S\sim\mathbb{P}_z&=\mathcal{S}\subset\mathbb{R}^{d},
\end{align}
where the spherical distribution samples vector $z_N$ from normal distribution and computes $z_S=\frac{z_N}{||z_N||_{2}}$. Prior works suggest that using uniform-like distributions ($\mathcal{U}$ and $\mathcal{S}$) is superior to a normal distribution ($\mathcal{N}$) in learning a general prior for human poses \cite{davydov2022adversarial}.
In this work, the generator is not intended to acquire a general prior pose knowledge, but rather to understand the failure modes of the pose estimator $\mathcal{P}$. 
Uniform-like priors tend to uniformly sample from the plausible poses while the normal distribution is mean-centered and can better find specific modes of the data. Therefore, we argue that normal distribution is a better case for our objective.

% In \textit{scenario 2}, the generator minimizes the loss of $\mathcal{P}$ on the generated data, and the feedback is defined as:
% \begin{equation}        
% F=\frac{1}{N}\sum_{i=1}^{N}||X_i-\hat{X}_i||_2,
% \end{equation}
% where $X$ and $\hat{X}$ are the ground-truth and estimated 3D poses and $N$ is the number of samples. In \textit{scenario 1}, the generator maximizes the loss of $\mathcal{P}$ on the generated data until it gets to a threshold, $c$. Therefore, the feedback is:
% \begin{equation}
% F=c-\frac{1}{N}\sum_{i=1}^{N}||X_i-\hat{X}_i||_2.
% \end{equation}

In \textit{Scenario 2}, the generator aims to minimize the loss of $\mathcal{P}$ on the generated data. The feedback in this scenario is defined as:
\begin{equation}        
\mathcal{L}_{fb}=\frac{1}{N}\frac{1}{J}\sum_{j=1}^{N}\sum_{i=1}^{J}||X_{i,j}-\hat{X}_{i,j}||_2,
\end{equation}
where $X$ represents the ground-truth 3D poses, $\hat{X}$ denotes the estimated 3D poses, $J$ is the total number of joints, and $N$ is the number of samples.
In \textit{Scenario 1}, the generator strives to maximize the loss of $\mathcal{P}$ on the generated data until it reaches a certain threshold, denoted by $c$. Consequently, the feedback in this scenario is given by:
\begin{equation}
\mathcal{L}_{fb}=c-\frac{1}{N}\frac{1}{J}\sum_{j=1}^{N}\sum_{i=1}^{J}||X_{i,j}-\hat{X}_{i,j}||_2.
\end{equation}

The overall objective function of the generator is to minimize the weighted summation of adversarial loss and feedback loss. In \textit{scenario 2}, the feedback loss enforces the generator to explore failure modes of $\mathcal{P}$ while the adversarial loss takes care of generated poses being plausible. On the other hand, in \textit{scenario 1} the generator tries to generate novel poses that do not significantly deviate from the original distribution of source data.  The overall loss of $\mathcal{G}$ is

\begin{equation}
    \mathcal{L}_{\text{$\mathcal{G}$}}= w_1\mathcal{L}_{adv}+w_2\mathcal{L}_{fb},
\end{equation}
where $\mathcal{L}_{adv}$ represents the least square GAN loss used for training the generator:
\begin{equation}
    \mathcal{L}_{adv}= \mathbb{E}_{z \sim \mathbb{P}_z}[ (\mathcal{D}(\mathcal{G}(z))-1)^2].
\end{equation}

The pose discriminator splits the human body into 6 parts including the torso, left/right leg, and left/right arm and head. 
The discriminator tries to distinguish real 3D poses from AMASS and synthetic 3D poses from the generator by taking into account the 6 body parts as well as the whole body parts. We use axis-angle joint angles $\theta$ as input for the discriminator $\mathcal{D}$. AMASS includes archives of human poses and we assume that using AMASS as the prior for the discriminator does not enforce the generated data to a specific sub-mode of human poses. In the appendix, we show the distribution of AMASS and the distribution of body poses in publicly available datasets such as 3DPWS (test-set). AMASS covers all models of the 3DPW test set and qualitatively proves that using AMASS is not problematic.

The discriminator $\mathcal{D}$ only enforces generated data in terms of body poses $\theta$ and does not take into account camera viewpoint $K$. The camera viewpoint is mainly affected by feedback from $\mathcal{P}$. The adversarial objective of the discriminator is:

\begin{equation}
\mathcal{L}_{\mathcal{D}}=\mathbb{E}_{\theta \sim \mathbb{P}_{\theta}}[( \mathcal{D}(\theta)-1)^2]+\mathbb{E}_{z \sim \mathbb{P}_z}[ \mathcal{D}(\mathcal{G}(z))^2]
\end{equation}
where $\mathbb{P}_\theta$ is the real pose distribution of the AMASS dataset.
% \begin{figure}[t]
% \begin{center}
% % \fbox{\rule{0pt}{2in} \rule{0.9\linewidth}{0pt}}
%    \includegraphics[width=0.8\linewidth]{AMSS_vs_3DPW.jpg}
% \end{center}
%    \caption{ Distribution of shoulder and hip joints from AMASS and 3DPW. AMASS provides a general distribution and therefore is suitable for learning a prior pose distribution. }
% \label{fig:AMASS}
% \end{figure}

\subsection{Animatable NeRF}
We use animatable human NeRF \cite{a-nerf} (A-NeRF) to render human images in new poses and from new viewpoints. A-Nerf enables rendering a human body in unseen poses and unseen viewpoints. The merit of A-NeRF compared with classical rendering methods is that it does not require human 3D scans and enables rendering personalized human images (with A-NeRF trained on personalized data). 

Given a sequence of frames from a person $[\textbf{I}_{k}]_{k=1}^{N}$, A-NeRF is aimed to optimize 3D poses $[\theta_{k}]_{k=1}^{N}$ and a parameterized body model $C_{\phi}$. $\phi$ and $\theta$ are optimized for an image reconstruction objective as follows:
\begin{equation}
   \mathcal{L}=\Sigma||C_{\phi}(\theta_{k})-I_{k}||_{1}+\lambda_{\theta}d(\theta_k-\hat{\theta}_{k})+\lambda_{t}||\frac{\partial^{2}{\theta}}{\partial{t}}||.
\end{equation}
The last term applies a smoothness prior and the middle term enforces the optimized $\theta$ to be close to $\hat{\theta}$ estimated by a 3D pose estimator. We render synthetic human images via ray marching as follows:

\begin{equation}
    C(u,v;\theta_{k})=\Sigma_{i=1}^{Q}T_{i}(1-exp(-\sigma_{i}\delta{i}))c_{i}, 
\end{equation}
where $(u,v)$ are 2d location in the image and $T$ is defined as $T_{i}=exp(-\sum_{j=1}^{i-1}\sigma_j \delta_i)$. $\sigma_j$ is the volume density at sample location $j$ along the ray and $\delta_i$ is the distance between two adjacent points along the ray. 

\subsection{Pose Estimator}
The pose estimator is trained with the image and 3D pose pairs $(\hat{I},\theta)$. The $\hat{I}$ is rendered by A-NeRF and $\theta$ is generated by $\mathcal{G}$. We assume that $\theta$ would be the ground truth 3D poses of the rendered image. However, the A-NeRF model is not perfect and there might be some errors in the rendered image specifically for complicated poses from a novel camera viewpoint. Therefore, we add a simple constraint on the loss of $\mathcal{P}$ to exclude samples with large errors. The $\mathcal{L}_\theta$ used for training the pose estimator is $\mathcal{L}_\theta=f(
||\theta_i-\hat{\theta}_i||_2)$ where $\hat{\theta}$ is the estimated pose and $f$ is:

\begin{equation}
  f(w) =
    \begin{cases}
      w & \text{if  } w<d\\
      0 & \text{otherwise}
    \end{cases}.      
\end{equation}
In the above formula, $d$ is a threshold to exclude samples with large errors. For the generalizability of our method, we use the same loss for fine-tuning any pose estimator. The additional losses used for prior working during their pre-training process are not used for fine-tuning. 
% \textbf{HybrIK}

% \textbf{SPIN}

\section{Experiments}

\textbf{Datasets.} We use three datasets that are not used during pre-training of pose estimator $\mathcal{P}$, including 3DPW, AGORA, and SKI-Pose, for evaluation. To further evaluate the effectiveness of the proposed \emph{PoseGen} on boosting the performance of $\mathcal{P}$ on IND datasets used in the pre-training procedure, we also perform an evaluation on 3DHP. Below we give the details of each dataset.
\begin{itemize}
    \item \textbf{3DPW} \cite{vonMarcard2018} includes in-the-wild images of two subjects performing different tasks including climbing, boxing, and playing basketball. We use the test set of 3DPW for evaluation.   
    \item \textbf{AGORA} includes realistic high-quality synthetic data from more than 150 subjects with varied clothing and with complex realistic backgrounds. AGORA includes frequent occluded images that make a unique dataset for evaluation of the generalizability of pre-trained models.
    \item \textbf{SKI-Pose} \cite{skipose} is captured in a ski resort from 5 professional athletes. SKI-Pose includes camera viewpoint and poses rarely seen in the training of pose estimators. We use the test set of SKI-Pose for evaluation. 
    \item \textbf{MPI-INF-3DHP (3DHP)} includes data from 8 subjects. The test set includes data from two of the subjects and includes in-the-wild and in-the-lab data. The training set of 3DHP has been used for training and the test set is used for evaluation as IND data.
\end{itemize}

\textbf{Pose Estimators.}
Our framework can be used to fine-tune any pre-trained pose estimator. We choose two popular pre-trained 3D human pose and mesh estimator models, namely HybrIK \cite{li2021hybrik} and SPIN \cite{kolotouros2019spin}. SPIN is a famous pre-trained model widely used as a baseline in recent works \cite{liu2023poseexaminer} and HybrIK is a recent method that has specifically shown promising results on cross-dataset evaluations.  HybrIK is pre-trained on 3DHP, Human3.6M \cite{h36m_pami}, and MSCOCO \cite{lin2015microsoft}. SPIN has been trained on Human3.6M, 3DHP, and LSP \cite{johnson2010clustered}.  We fine-tuned these pre-trained models with our framework and evaluated them on unseen datasets.

\textbf{NeRF Model.} The NeRF model \cite{a-nerf} has been trained on 1500 synthetic 3D poses from \cite{CMU}. The 3D poses were rendered by \cite{Varol_2017_CVPR} from 9 different camera viewpoints. The total number of images in the training set was 10800 $512 \times 512$ images. We keep the NeRF model frozen during training and data generation. 

\textbf{Evaluation Metrics.}
 Following previous work, we use mean-per-joint position error (MPJPE) and mean-per-joint position error after Procrustes alignment (PA-MPJPE) with the ground truth 3D poses. We also report PCK on the 3DHP dataset.
 
\subsection{Quantitative Results} Tables \ref{tab:AGORA}, \ref{tab:3DPW}, and \ref{tab:SKI} show the evaluation results of \emph{PoseGen} on AGORA, 3DPW, and SKI-Pose under scenario 2 assumptions. In the upper section of the tables, we show the results of pose estimators after fine-tuning (FT) on the designated dataset. In the lower section of the tables, we show the evaluation results of the pre-trained model (PT) on the test set of the designated dataset. The difference between the results of FT and PT models indicates how much the dataset is OOD for the PT models. The difference between FT and PT on AGORA, 3DPW, and SKI-Pose are about 61 mm, 17mm, and 73mm, respectively. Therefore, the test-set of AGORA and SKI-Pose are highly OOD for PT HybrIK. We show the results of SPIN and HybrIK after fine-tuning with \emph{PoseGen} in the last part of Tables \ref{tab:AGORA}-\ref{tab:SKI}. \emph{PoseGen} relatively improves HybrIK for about 4\% and 5\% in terms of MPJPE on AGORA and 3DPW, respectively. The improvements of SPIN are 15\% and 7\% on AGORA and 3DPW, respectively. 

Table \ref{tab:3DHP} shows the results of our method on the 3DHP dataset. Since 3DHP has been used for pre-training of HybrIK and SPIN, we consider that as an IND dataset. \emph{PoseGen} improves SPIN and HybrIK for about 3\% in terms of MPJPE. Therefore, our method is also effective in improving the baseline pose estimators on IND datasets. Although the 3DHP dataset has already been used during pre-training, the test set of 3DHP is challenging and involves novel poses. Therefore, pre-trained HybrIK obtains an MPJPE of 99.3 mm on 3DHP while it obtains an MPJPE of 88.7 mm on the 3DPW dataset. 

Table \ref{tab:SKI} shows the results on the SKI-Pose dataset. The definition of joints (e.g. hip and shoulders) and bone lengths of ground truth 3D poses of the SKI dataset are a little bit different from SMPL joint definitions. Moreover, most of the poses of SKI-Pose contain athletes bent toward the side and from novel viewpoints. Therefore, the errors reported on SKI-pose are greater compared with other datasets. However, \emph{PoseGen} still improves HybrIK and SPIN on the SKI-Pose dataset. Comparing all four benchmarks, our method is effective in improving baseline models.

\begin{figure}[t]
\begin{center}
   \includegraphics[width=1\linewidth]{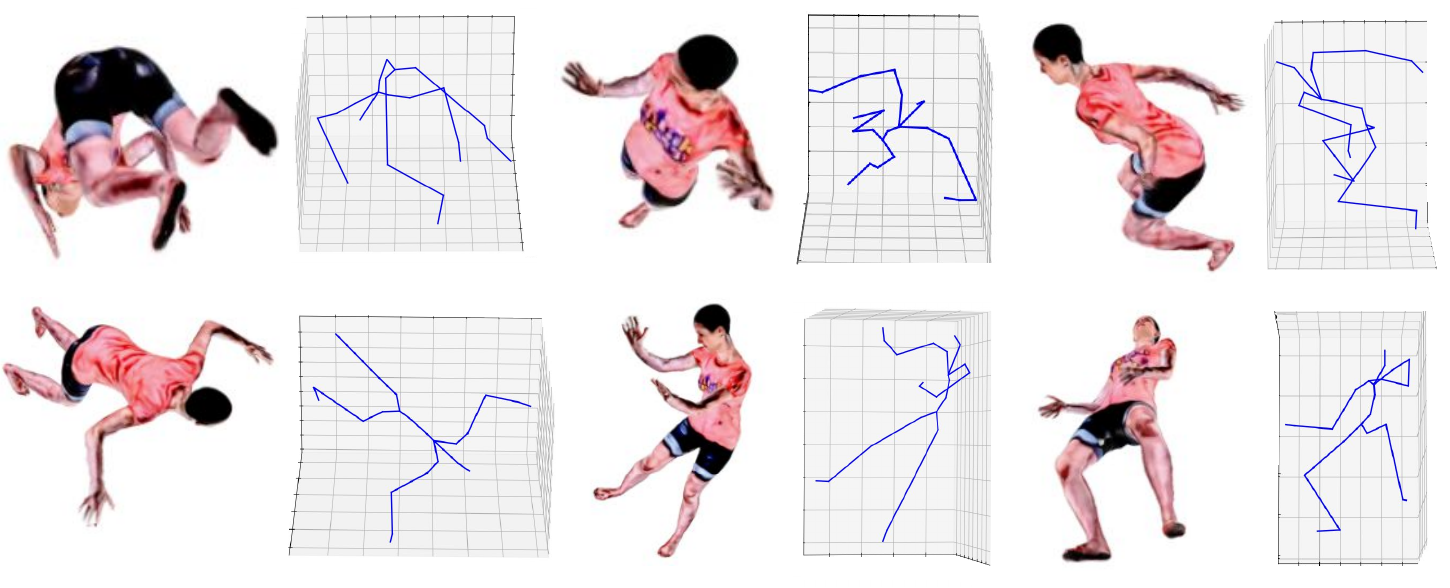}
\end{center}
   \caption{ Sample images and 3D poses generated by \emph{PoseGen}. Images are from novel poses and novel camera viewpoints. The 3D poses and images qualitatively are good, even though images are rendered from rare viewpoints. }
\label{fig:qualitative}
\end{figure}

\begin{table}
\small
\centering
\caption{Results on AGORA dataset. Models that are trained on the training set of AGORA are shown with a checkmark.}
\label{tab:AGORA}
\begin{tabular}{ p{3.8cm}|c|cc}
\hline
Method&FT&MPJPE$\downarrow$&NMJE$\downarrow$\\
\hline
SPIN (Kolotouros et al. 2019)&\checkmark&153.4&199.2\\
BEV\cite{sun2022putting} &\checkmark&105.3&113.2\\
CLIFF \cite{li2022cliff}&\checkmark&81.0&89.0\\
HybrIK \cite{li2021hybrik}&\checkmark&77.0&84.6\\
\hline
SPIN(Kolotouros et al. 2019)&&175.1&223.1\\
EFT (Joo et al. 2021) &&165.4&203.6\\
VIBE(Kocabas et al. 2020) &&146.2&174.0 \\
HybrIK\cite{li2021hybrik}&&137.9&166.1\\\hline
Ours+SPIN&&\bf{158.9\scriptsize{(-16.2)}}&\bf{189.2\scriptsize{(-34)}}\\
% Ours+HybrIK&&\textbf{48.4(-0.9)}&\textbf{84.5(-4.2)}\\
Ours+HybrIK&&\bf{132.7\scriptsize{(-5.2)}}&\bf{159.9\scriptsize{(-6.2)}}\\
% Ours+CLIFF&&\\
\hline
\end{tabular}
\end{table}

\begin{table}
\small
\centering
\caption{Results on 3DPW dataset. Models that are trained on the training set of 3DPW are shown with a checkmark.}
\label{tab:3DPW}
\begin{tabular}{ p{3.7cm}|c|cc}
\hline
Method&FT&PA-MPJPE$\downarrow$&MPJPE$\downarrow$ \\
\hline
EFT (Joo et al. 2021) &\checkmark&55.7&--\\
VIBE (Kocabas et al. 2020)&\checkmark&51.9&82.9\\
HybrIK \cite{li2021hybrik}&\checkmark&41.8&71.3\\
CLIFF \cite{li2022cliff}&\checkmark&43.0&69.3\\
\hline
% Sim2real\cite{Sim2real} &&74.7&--\\
% Zhang \textit{et al.}\cite{zhang2020inference}&&70.8&--\\
\cite{Li_Pun_2023}&&76.8&-\\
SPIN(Kolotouros et al. 2019)& &59.2&96.9\\
PoseAug (Gong et al. 2021)&&58.5&94.1\\
VIBE (Kocabas et al. 2020) &&56.5&93.5\\
\cite{choi2022learning} &&51.5&93.5\\
HybrIK \cite{li2021hybrik}&&49.3&88.7\\
% CLIFF&&46.4&73.9\\ 
\hline
Ours+SPIN&&\textbf{56.2\scriptsize{(-3.0)}}&\textbf{89.7\scriptsize{(-7.2)}}\\
% Ours+HybrIK&&\textbf{48.4(-0.9)}&\textbf{84.5(-4.2)}\\
Ours+HybrIK&&\textbf{48.3\scriptsize{(-1.0)}}&\textbf{84.4\scriptsize{(-4.3)}}\\
% Ours+CLIFF&&46.7&\textbf{72.7(-1.2)}\\
\hline
\end{tabular}
\end{table}

\begin{table}
\small
\centering
\caption{Results on SKI-Pose dataset. * Trained using multi-view cameras. ** Trained using multi-view cameras and partial 3D annotations.}
\label{tab:SKI}
\begin{tabular}{ p{3.7cm}|c|cc}
\hline
Method&FT&PA-MPJPE$\downarrow$&MPJPE$\downarrow$\\
\hline
\cite{Rhodin_2018_CVPR}&\checkmark&-&85\\
\cite{Wandt_2021_CVPR}**&\checkmark&89.6&128.1\\ 
SPIN(Kolotouros et al. 2019)&\checkmark&57.2&94.3\\ \hline
% SPIN&\checkmark&&\\
% HybrIK \textit{et al.}&\checkmark&&\\ \hline
SPIN(Kolotouros et al. 2019)&&135.5&288.9\\
HybrIK \cite{li2021hybrik}&&125.5&205.2\\ \hline
Ours+SPIN&&\bf{130.6\scriptsize{(-5)}}&\bf{250.9\scriptsize{(-33)}}\\
% Ours+HybrIK&\checkmark&\textbf{65.3(-1.5)}&\textbf{96.6(-2.7)}\\
Ours+HybrIK&&\textbf{124.5\scriptsize{(-1)}}&\textbf{204.2\scriptsize{(-1)}}\\
% Ours+CLIFF&&\\
\hline
\end{tabular}
\end{table}
\begin{table}
\small
\centering
\caption{Results on 3DHP dataset. All models use the 3DHP dataset for training. }
\label{tab:3DHP}
\begin{tabular}{ p{3.7cm}|c|cc}
\hline
Method&FT&PCK$\uparrow$&MPJPE$\downarrow$\\
\hline
HMR \cite{kanazawa2018endtoend}&\checkmark&72.9&124.2\\
SPIN(Kolotouros et al. 2019)&\checkmark&76.4&105.2\\
HybrIK \cite{li2021hybrik}&\checkmark&80.0&99.3\\ \hline
Ours+SPIN&\checkmark&\bf{80.9\scriptsize{(+4.5)}}&\bf{101.7\scriptsize{(-3.5)}}\\
% Ours+HybrIK&\checkmark&\textbf{65.3(-1.5)}&\textbf{96.6(-2.7)}\\
Ours+HybrIK&\checkmark&\textbf{85.0\scriptsize{(+1.0)}}&\textbf{96.7\scriptsize{(-2.6)}}\\
% Ours+CLIFF&&\\
\hline
\end{tabular}
\end{table}

\begin{table}
\centering
\small
\caption{Ablation study on the main components of PoseGen.}
\label{tab:ablate}
\begin{tabular}{ c|ccc|cc}
\hline
&$\theta$&$K$&FB&PA-MPJPE&MPJPE\\
\hline
Baseline &&&&59.2&96.9\\
A1 &\checkmark&&&57.1&95.8\\
A2 &\checkmark&\checkmark&&56.2&92.7\\
A3 &\checkmark&\checkmark&\checkmark&55.8&91.3\\
\hline
\end{tabular}
\end{table}

\begin{table}
\centering
\small
\caption{Ablation study on scenarios 1 and 2.}
\label{tab:ablate_scenario}
\begin{tabular}{ c|cc}
\hline
&PA-MPJPE&MPJPE\\
\hline
Scenario 1 &55.9&92.6\\
Scenario 2 &56.0&91.8\\
\hline
\end{tabular}
\end{table}

\begin{table}
\centering
\small
\caption{Ablation study on prior distributions.}
\label{tab:ablate_dist}
\begin{tabular}{ c|cc}
\hline
Distribution&PA-MPJPE&MPJPE\\
\hline
Uniform ($\mathcal{U}$) &55.9&91.9\\
Spherical ($\mathcal{S}$) &56.2&92.4\\
Normal ($\mathcal{N}$) &\textbf{55.8}&\textbf{91.3}\\
\hline
\end{tabular}
\end{table}

\begin{figure}[!h]
\begin{center}
   \includegraphics[width=1\linewidth]{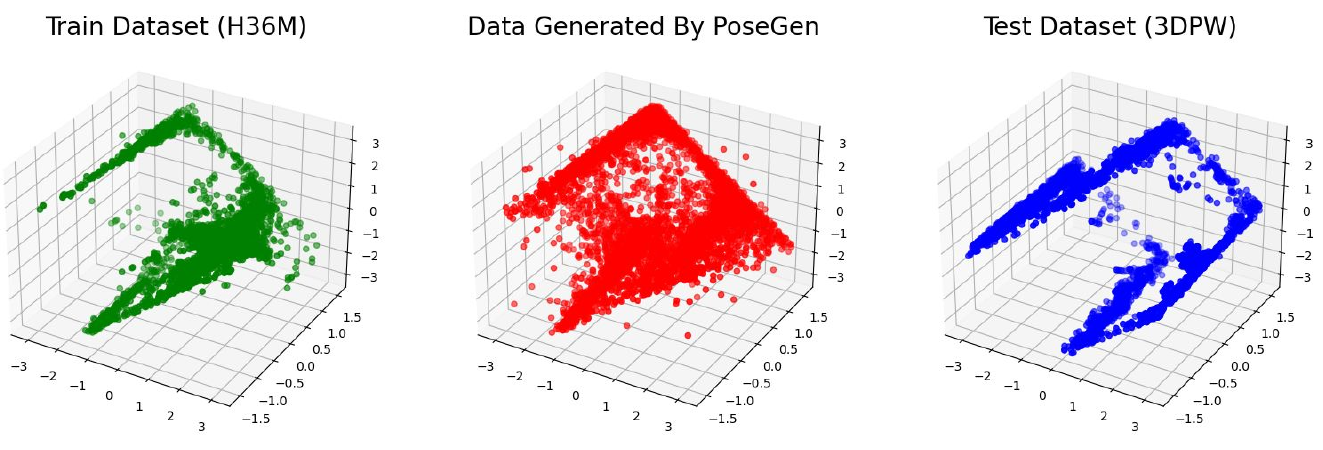}
\end{center}
   \caption{The distribution of camera viewpoint of the train dataset (H3.6M), the test dataset (3DPW), and the synthetic data generated by \emph{PoseGen}.}
\label{fig:distribution}
\end{figure}

\begin{figure}[h]
\begin{center}
   \includegraphics[width=1\linewidth]{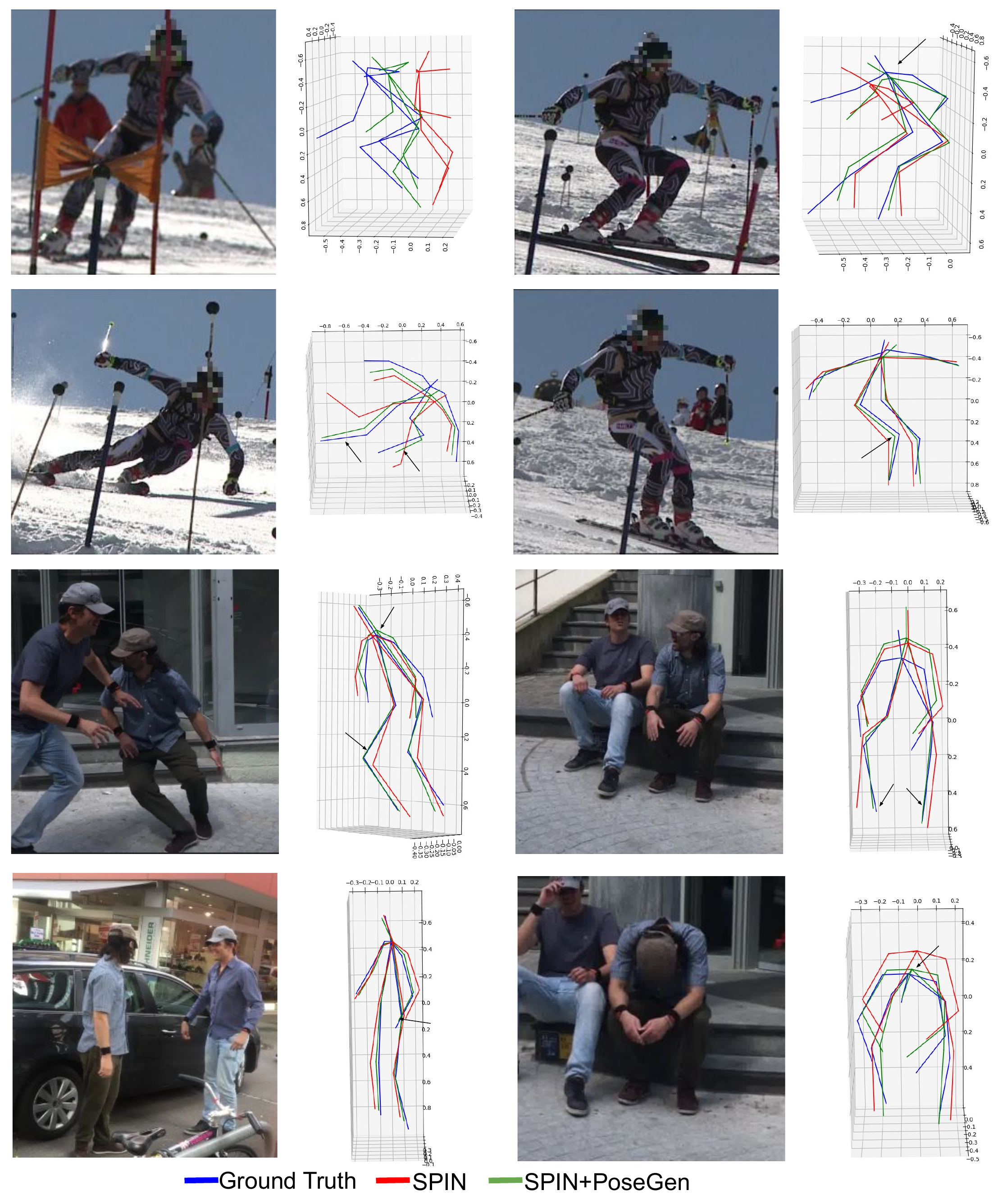}
\end{center}
   \caption{Predictions of SPIN and \emph{PoseGen(+SPIN)} on in-the-wild images from SKI-Pose and 3DPW. On the challenging SKI-Pose dataset, \emph{PoseGen} improves the predictions in the z-direction (depth) and global rotation of the human body.}
\label{fig:qualitative2}
\end{figure}

\begin{figure}[h]
\begin{center}
   \includegraphics[width=1\linewidth]{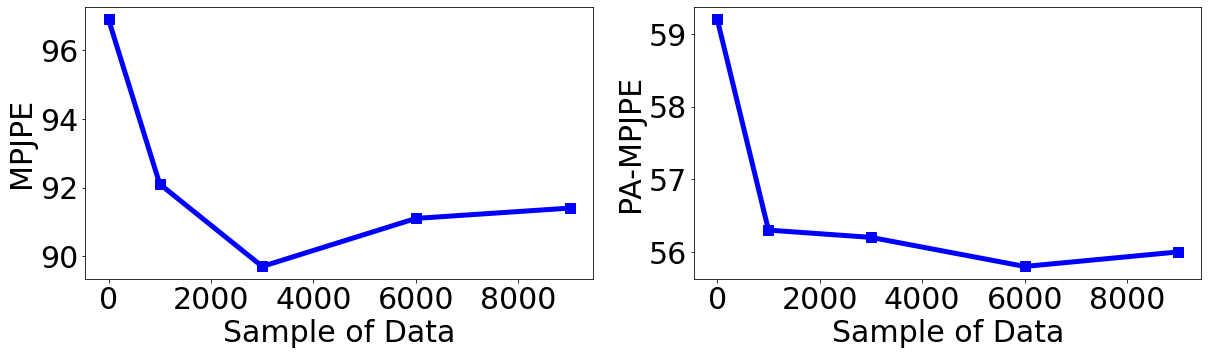}
\end{center}
   \caption{Performance improvement while increasing the number of generated samples.}
\label{fig:samples}
\end{figure}

\subsection{Qualitative Results}
Figure \ref{fig:qualitative} shows some samples of generated data by \emph{PoseGen} and their 3D poses. Images are from novel camera viewpoints that are not present in the datasets used for pre-training of the pose estimators. Most of the public datasets only have chest-view cameras. Our generated data finds failure modes on top-view and bottom-view cameras and has generated such samples. Among the 4 benchmarks, the test set of AGORA is the only benchmark that covers top-view cameras. Comparing 3D poses and rendered images shows that rendered images follow the input 3D poses. Figure \ref{fig:distribution} shows the distribution of camera viewpoints in the training dataset (H3.6M), test dataset (3DPW), and synthetic data generated by our method. ur framework generates samples from unseen viewpoints, thus covering OOD viewpoints. Figure \ref{fig:qualitative2} shows the predictions of SPIN and PoseGen+SPIN vs. ground truth 3D poses on images from SKI-Pose and 3DPW. The top row shows that \emph{PoseGen} improves SPIN in terms of the global orientation. Moreover, the joint angles on novel poses of athletes in the ski resort are better predicted.

\subsection{Ablation Studies}
\textbf{Components of \emph{PoseGen}.} Table \ref{tab:ablate} shows the results of \emph{PoseGen} after excluding the main components of the framework. In \textit{A1} we only generate novel poses and render poses which improved the baseline for 1.1 mm (MPJPE). In \textit {A2} we generate both novel poses and camera-viewpoint that further improves {A1} for 3 mm. Adding feedback in \textit{A3} improves \textit{A2} for 1mm. The ablation study shows that all components are critical in improving the baseline models. Moreover, generating data from novel camera viewpoints has a major impact on improving the robustness of pre-trained models. This is well-aligned with the findings of prior work that generate 2D-3D pose pairs \cite{Gholami_2022_CVPR}.  

\textbf{Scenarios.} Table \ref{tab:ablate_scenario} compares the results of \textit{scenario 1} and \textit{scenario 2} on 3DPW. In \textit{scenario 2} and \textit{scenario 1} we obtained an MPJPE of 91.8 and 92.6, respectively. These results show that generating OOD samples in \textit{scenario 2} makes the pre-trained model more robust on unseen OOD samples. 

\textbf{Prior Distributions.} Table \ref{tab:ablate_dist} shows the ablation study on different prior distributions. Previous work has shown that uniform and spherical distributions are more effective in learning a general prior for the human pose. Our experiments show that normal distribution is a better choice for our framework. Generating data with normal, uniform, and spherical distribution results in MPJPE of 91.3, 91.9, and 92.4, respectively on the 3DPW dataset. We hypothesize that our method is aimed at finding the failure modes of the pre-trained model. In contrast to prior works that try to find a smooth uniform prior, failure modes are usually discontinuous. Therefore, having a uniform distribution ($\mathcal{U}$ and $\mathcal{S}$) is not a proper prior for \emph{PoseGen}.

\textbf{Dataset Size.} We increased the number of generated samples from 1K to 9k, and our experiments showed in Figure 6 that we could obtain the best performance with only 6K samples. \emph{PoseGen} improves the performance of SPIN from  59.2 mm to 55.8 mm in terms of PA-MPJPE with only 6K samples. AGORA \cite{agora} improves the performance of SPIN on the 3DPW dataset to 55.8 mm by generating a dataset of 14K images (each including multiple subjects) with more than 350 subjects. Therefore, our method is more efficient compared with competitors.

\subsection{Limitations}
\begin{figure}[t]
\begin{center}
   \includegraphics[width=1\linewidth]{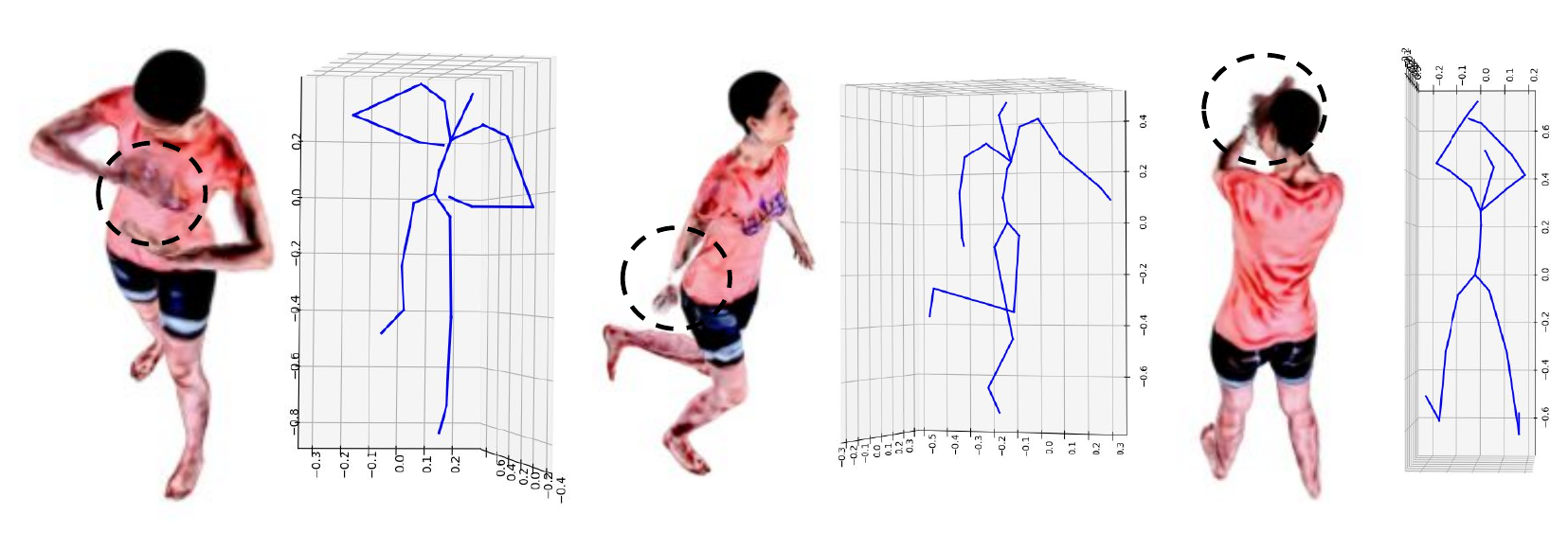}

   \caption{ Failure models of \emph{PoseGen}. A-NeRF is not perfect in rendering images from the novel viewpoint and novel poses. }
\label{fig:failure}
\end{center}
\end{figure}

The NeRF model used in this study has some limitations in rendering complex poses. Figure \ref{fig:failure} shows some failure cases in rendering images from novel poses. The A-NeRF model is only capable of rendering images for a single subject. In order to render images for different subjects, we need to use separate NeRF checkpoints trained on that specific subject data. Future work should extend the experiments by generating a dataset that includes different subjects. 
% Although we only used one character in generating data, we could improve the baseline models significantly. 
We expect that having more subjects in the framework will further improve the performance. 
% We neglected human body shape as another dimension in generating data, due to the limitation of A-NeRF.
% Future work should extend NeRF models to be able to render poses with varied shapes. 
% Though we focused on generating data for single-person pose estimation, \emph{PoseGen} can be extended to generate multi-person datasets.

\section{Conclusion}
In this work, we proposed an end-to-end framework for generating a 3D human pose dataset using NeRF. Our experiments showed that NeRFs are capable of generating datasets to improve the robustness of the pre-trained model. NeRFs can be trained on use-specific images and therefore the proposed framework can be used in future work to generate user-specific datasets. We performed experiments on two scenarios where we generated data 1) to minimize, and 2) to maximize the loss of a pre-trained model. We showed that the second scenario results in better performances. Moreover, we performed experiments on the prior distributions and showed that uniform and spherical prior distributions are not appropriate for the specific objective of this work.

\newpage

\section{Appendix: Experiment on A User-Specific Dataset}
\begin{figure}[h]
\begin{center}
   \includegraphics[width=1\linewidth]{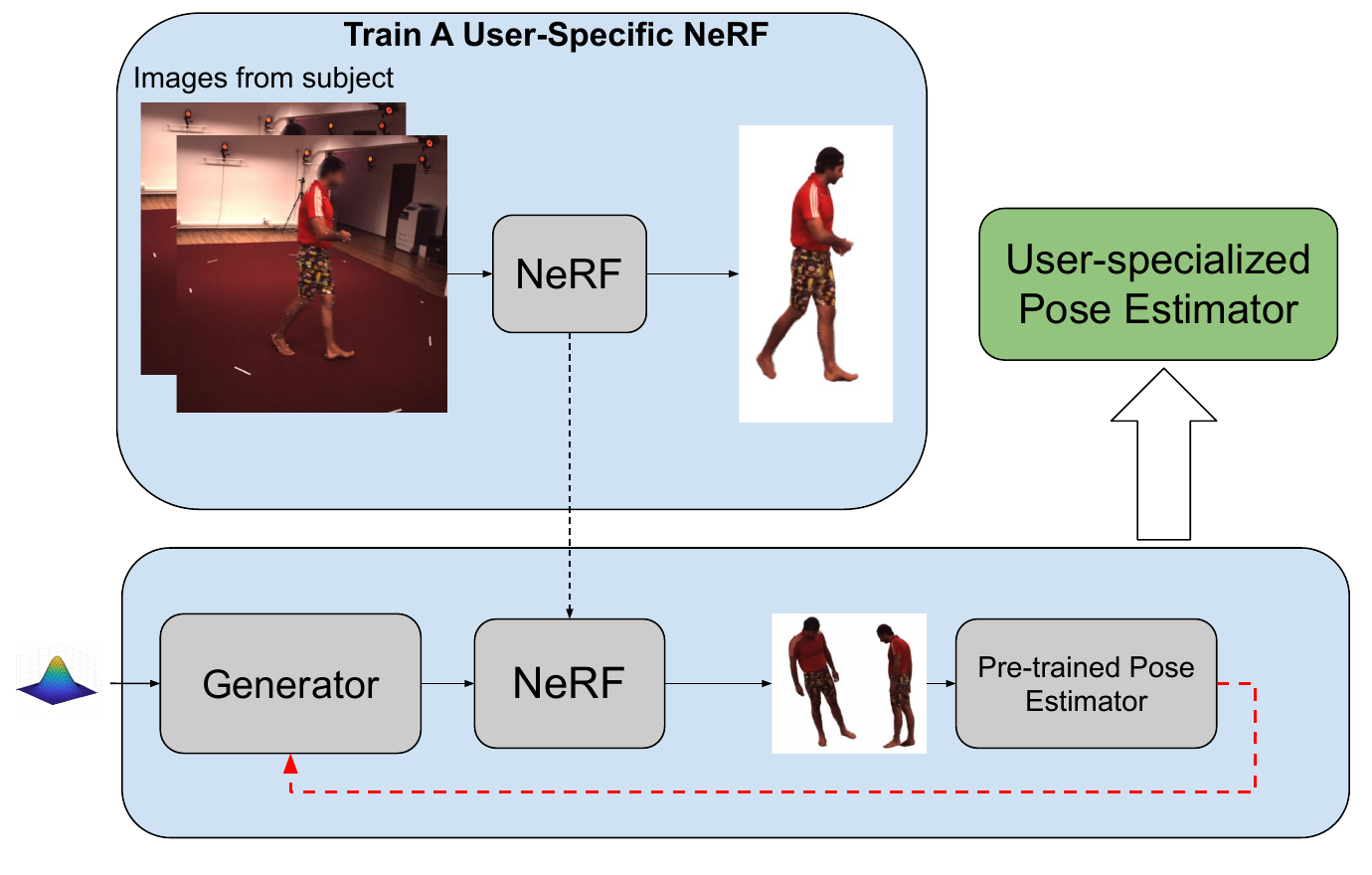}
\end{center}
   \caption{ Overview of the steps required for preparing user-specialized pose estimator. Given sample videos from the subject, A-NeRF has been trained and then \emph{PoseGen} fine-tunes the pre-trained model.}
\label{fig:usd}
\end{figure}

The proposed method is capable of preparing user-specific 3D human pose estimators. In order to prepare such a model, we only need samples of videos from the subject to train a NeRF model specialized for that subject. Then, the trained NeRF model can be used to generate OOD samples for the pre-trained model via \emph{PoseGen}. Figure \ref{fig:usd} shows an overview of the steps required for preparing user-specialized models. The proposed framework requires little input from the user and can return a model that is more accurate than public pre-trained models for that subject. To the best of our knowledge, this is the first work that proposes a flexible framework for preparing such models. 

\begin{figure*}[h]
\begin{center}
   \includegraphics[width=0.9\linewidth]{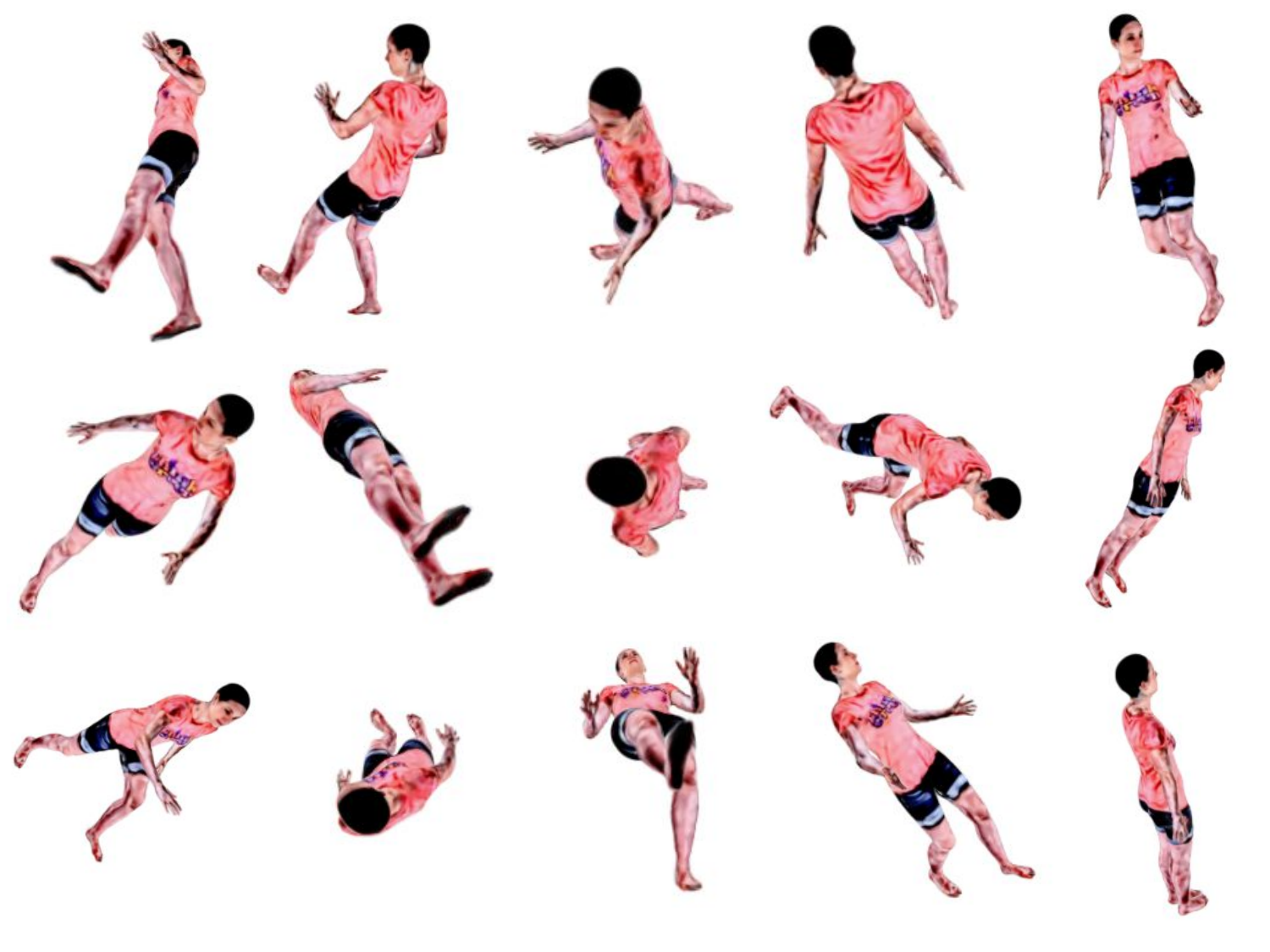}
\end{center}
   \caption{ Samples of the user-specific test dataset (USD). Images are rendered given ground truth 3D poses from the test set of 3DPW and from random viewpoints.}
\label{fig:testset}
\end{figure*}

In order to examine this scenario, we create a user-specific test dataset (USD). We sample 1K poses from the test set of the 3DPW dataset and render images from random camera viewpoints. We use a SURREAL dataset character to render images that we name $S0$ for simplicity. Figure \ref{fig:testset} shows some samples of the USD. Since camera viewpoints are sampled randomly from a uniform distribution, USD is more challenging compared with the test of 3DPW. All samples of USD are images of $S0$ in different poses and from variant viewpoints. A NeRF model is also trained on separate videos of $S0$. The videos that were used for training of A-NeRF are not in USD. The trained A-NeRF checkpoint is then used via \emph{PoseGen} to prepare the user-specialized model. SPIN and HybrIK are fine-tuned with \emph{PoseGen} on the OOD samples generated by \emph{PoseGen}.

\begin{table}[h]
\centering
\caption{Results on the user-specific dataset.}
\label{tab:testset}
\begin{tabular}{ p{3cm}|c|ll}
\toprule
Method&FT&PA-MPJPE$\downarrow$&MPJPE$\downarrow$\\
\hline
% SPIN (Kolotouros et al. 2019)&\checkmark&&\\
% HybrIK \cite{li2021hybrik}&\checkmark&&\\
% \midrule
SPIN &&93.8&209.2\\
HybrIK &&92.9&156.9\\
\midrule
Ours+SPIN&&\textbf{69.0 \scriptsize{(-24.8)}}&\textbf{98.2  \scriptsize{(-111)}}\\
% Ours+HybrIK&&\textbf{48.4(-0.9)}&\textbf{84.5(-4.2)}\\
Ours+HybrIK&&\textbf{69.2 \scriptsize{(-23.7)}}&\textbf{94.7  \scriptsize{(-62.2)}}\\
\bottomrule
\end{tabular}
\end{table}

\begin{table*}[!h]
\centering
\small
\caption{Results on 3DPW dataset. Models trained on the training set of 3DPW are shown with a checkmark.}
\label{tab:3DPW}
\begin{tabular}{ p{8cm}|c|cc}
\toprule
Method&FT&PA-MPJPE$\downarrow$&MPJPE$\downarrow$\\
\hline
EFT (Joo et al. 2021) &\checkmark&55.7&--\\
VIBE (Kocabas et al. 2020)&\checkmark&51.9&82.9\\
HybrIK \cite{li2021hybrik}&\checkmark&41.8&71.3\\
CLIFF \cite{li2022cliff}&\checkmark&43.0&69.3\\
\midrule
% Sim2real\cite{Sim2real} &&74.7&--\\
% Zhang \textit{et al.}\cite{zhang2020inference}&&70.8&--\\
\cite{Li_Pun_2023}&&76.8&-\\
SPIN (Kolotouros et al. 2019)& &59.2&96.9\\
PoseAug (Gong et al. 2021)&&58.5&94.1\\
VIBE (Kocabas et al. 2020) &&56.5&93.5\\
\cite{choi2022learning} &&51.5&93.5\\
\cite{yang2023capturing}&&49.7&90.0\\
ImpHMR \cite{cho2023implicit}&&49.8&81.8\\
HybrIK \cite{li2021hybrik}&&49.3&88.7\\
PoseExaminer \cite{liu2023poseexaminer}&&48.0&77.5\\
PyMAF-X \cite{pymafx2023}&&47.1&78.0\\
CLIFF \cite{li2022cliff}&&46.4&73.9\\ 
\midrule
Ours+SPIN&&56.2&89.7\\
% Ours+HybrIK&&\textbf{48.4(-0.9)}&\textbf{84.5(-4.2)}\\
Ours+HybrIK&&48.3&84.4\\
\textbf{Ours+CLIFF}&&\textbf{46.4}&\textbf{73.0}\\
\bottomrule
\end{tabular}
\end{table*}

We examine pre-trained SPIN and HybrIK on the USD and obtain an MPJPE of 209.2 mm and 156.9 mm. The obtained PA-MPJPE with SPIN and HybrIK are 93.8 and 92.9.  The results of the baseline models are worse on USD compared with 3DPW (PA-MPJPE: 93.8 vs. 59.2) since USD is more challenging than the 3DPW (test set). Finetuning the baseline models with \emph{PoseGen} significantly improves the results of both models. We obtain 25\% and 26\% relative improvements over SPIN and HybrIK. The proposed method is capable of preparing accurate models according to the needs of subjects. 

\section{Appendix: Experiment on an Additional Baseline}
In the main body, we used HybrIK and SPIN as two baselines and showed that the data generated with \emph{PoseGen} can improve these two baseline models with an average of 6\% relative improvements. In order to further evaluate the robustness of \emph{PoseGen}, we use \emph{PoseGen} to enhance the generalization of a better baseline model, CLIFF. pre-trained CLIFF obtains an MPJPE of 73.9 mm on the 3DPW dataset. Fine-tuning CLIFF with our framework improves the performance of CLIFF and results in state-of-the-art results compared with recent works. \emph{PoseGen}+CLIFF results in an MPJPE of 73.0 which is better than CLIFF with 0.9 mm. Table \ref{tab:3DPW} shows the results our results compared with prior arts. 

\newpage
\bibliography{aaai24}

\end{document}